\documentclass[10pt,journal,a4]{IEEEtran}

\usepackage{amssymb,amsthm,amsmath,amsbsy,epsfig,float,lscape,makeidx,bm}
\usepackage{multirow}
\usepackage{color,algorithm}
\usepackage[square,sort,comma,numbers]{natbib}
\usepackage[bottom]{footmisc}

\definecolor{MyDarkBlue}{rgb}{0,0.08,0.45}
\definecolor{yellow}{rgb}{0.99,0.99,0.70}
\definecolor{white}{rgb}{1.0,1.0,1.0}
\definecolor{black}{rgb}{0.00,0.00,0.00}
\definecolor{grey1}{rgb}{0.9,0.9,0.9}
\definecolor{grey2}{rgb}{0.8,0.8,0.8}

\def\argmax{\mathop{\rm argmax}}
\def\argmin{\mathop{\rm argmin}}

\newcommand{\real}{\ensuremath{\mathbb{R}}}

\def\sl{\mathop{\mathfrak s\mathfrak l}\nolimits}

\def\argmax{\mathop{\rm argmax}}
\def\argmin{\mathop{\rm argmin}}

\def \P {\mathcal{P}}

\newcommand{\cC}{\ensuremath{\mathbb{C}}}

\newtheorem{lemma}{Lemma}

\begin{document}
\tabcolsep 1pt
		\title{Analyzing Dynamical Brain Functional Connectivity As Trajectories on Space of Covariance Matrices
		 }
		\author{Mengyu Dai, Zhengwu~Zhang, and~Anuj~Srivastava}	
		
		\maketitle
		
			\begin{abstract}
			Human brain functional connectivity (FC) is often measured as the similarity of functional MRI responses across brain regions when a brain is 
			either resting or performing a task. 
			This paper aims to statistically analyze the dynamic nature of FC by representing the collective time-series 
			data, over a set of brain regions, as a trajectory on 
			the space of covariance matrices, or symmetric-positive definite matrices (SPDMs). 
			We use a recently developed metric on the space of SPDMs for quantifying differences across 
			FC observations, and for clustering and classification of FC trajectories. To facilitate large scale and high-dimensional data analysis, 
			we propose a novel, metric-based dimensionality reduction technique to reduce data from large SPDMs 
			to small SPDMs.  
			We illustrate this comprehensive framework using data from the Human Connectome Project (HCP) database for multiple subjects and tasks, 
			with task classification rates that match or outperform state-of-the-art techniques.  									
		\end{abstract}
		

\footnote{This work has been submitted to the IEEE for possible publication. Copyright may be transferred without notice, after which this version may no longer be accessible.}	

\section{Introduction}\label{sec:introduction}
There has been a great interest in studying brain functional connectome generated from functional MRI (fMRI), which measures the blood oxygen level dependent (BOLD) contrast signals inside the brain over a period of time.
The recent Human Connectome Project (HCP) \citep{hcp2013} facilitates such research by providing high-quality publicly available dataset.  The functional connectome, also referred to as functional connectivity (FC), is estimated  as a set of {\it statistical dependencies} of fMRI signals among remote anatomical regions. These dependencies are expressed as quantifications of similarity, {correlation}, or covariance between simultaneous BOLD measurements across regions in human brain. Recent studies \cite{Hutchison-etal:2013} have shown
that FC is a dynamic process and changes continuosly, even in the resting state.  The short-term FC is often represented as a correlation or covariance matrix of fMRI data over a small time window, with the matrix size equal to the number of brain regions being considered. 

Our primary goal is to quantitatively represent and compare dynamic FC between different anatomical regions, 
over long intervals representing either performances of certain tasks or 
with brain at resting state. We achieve this by representing dynamic FCs as trajectories on space of covariance matrices, and comparing these
trajectories using a Riemmanian metric on this space. This, in turn,  
leads to classification of dynamic FCs and prediction of various kinds of disease or stimuli using fMRI data. 
One challenge is the high dimensionality of the observed data - 
it becomes expensive to efficiently compare covariance trajectories when 
the dimension of covariances (i.e. the number of regions being studied) 
is very large. To handle this issue, we provide a metric-based, dimension-reduction method that helps reduce large covariance matrices to small covariance matrices, using a metric-based approach, and significantly improves computational efficiency of the ensuing statistical analysis.

A significant progress has been made over the last
few years in the field of fMRI data analysis. One of the foci has been the classification of 
task and external stimuli using fMRI data. In \cite{8a1148a3217b4fad86e6e0026e814524}, the authors used a multivariate pattern analysis (MVPA) approach to test the hypothesis that motivation-related enhancement of cognitive control results from improved encoding and representation of task set information. In the study presented in 
\cite{Sarraf070441}, fMRI data were studied using deep learning tools for Alzheimer's disease prediction. 
Barachant et al. \cite{Barachant:2013} studied covariance matrices classification using SVM classifiers under a Riemannian metric based kernel. Zhao et al. \cite{Zhao2018ARF} proposed a Riemannian framework for performing group analysis on longitudinal RS-fMRI data. Experiments studied the potential use of their framework in longitudinal connectivity analysis. Dodero et al. \cite{7493507} extracted traces of brain function by constructing trajectories in a low-dimensional space. Their approach allows studying dynamic FC in detail from the moment-to-moment fluctuations of the brain signals during resting state or other conditions. Qiu et al. \cite{QIU201552} proposed a framework for computing the mean of a set of brain functional networks and embedding brain functional networks into a low-dimensional space. They used Locally Linear Embedding (LLE) under the Log-Euclidean Riemannian metric space for this embedding. Dodero et al. \cite{7163812} presented a general computational framework for classifying connectivity graphs using kernels defined on the Riemannian manifold of positive definite matrices. Based on representation through a regularized graph Laplacian, this method can also be applied to classification of any kind of connectivity graphs with non-negative weights. Huang et al. \cite{2018arXiv181210050H} proposed a regression approach that smooths out the noise by exploiting the geometric structure of correlation matrices. They demonstrated that the method is useful for identifying distinct state spaces in the resting-state connectivities of males and females through the transition matrices of the common states. 
Despite these achievements, these methods either did not provide quantitive measures to represent and compare dynamic FCs at the 
network level, or did not have efficient solutions in region-based large-scale multi-class classification problems. 
Here we aim to address these issues using a novel combination of representing 
dynamic FC using a sequence of 
symmetric positive definite matrices (SPDMs), a Riemannian structure on SPDMs, and 
a tool for dimension reduction from large SPDMs to small SPDMs.

We first parcellate human brain into functional units or regions, and represent
short-term FC as SPDMs, with entries corresponding to 
covariances of localized  fMRI signals. We use a sliding window to segment multivariate time series into overlapping temporal blocks, and estimate a covariance matrix for each block. The original time series data (BOLD signal) can thus be converted into a 
time series of covariance matrices, termed as a covariance trajectory or SPDM trajectory. 
Next, we utilize a Riemannian structure (described in several places including \cite{zhang-etal:2018}) on the space of SPDMs to facilitate the comparison of trajectories. 
To improve computational efficiency, we propose a novel dimension-reduction approach based on the chosen Riemannian metric of SPDMs. The novel contributions of this paper are as follows:
\begin{enumerate}
	\item  {\bf Dynamic FC Representation Using SPDM Trajectories}: We represent the dynamic FCs as SPDM trajectories and compares these trajectories based on an advanced Riemannian framework. 
	
	\item {\bf Metric-Based Task Classification}: We use a Riemannian metric to perform registration, comparison, and classification of dynamic FCs as covariance trajectories. 
	
	\item {\bf SPDM Dimension Reduction}: We introduce a metric-based dimension-reduction technique to map large SPDMs into small ones, thus improving the computational efficiency and facilitating statistical analysis of data 
	involving hundreds of regions. 
\end{enumerate}

The rest of this paper is organized as follows. In Section II, we describe the mathematical components of the proposed framework, including: 
the estimation and representation of FC using SPDMs, Riemannian metrics for comparing SPDM trajectories, and dimension reduction for large SPDMs. Section III presents experimental results using both simulated and real HCP data. The paper ends with a short discussion and summary in Section IV. 

\section{Methodology}
In this section we lay out different components of our framework. 
After standard preprocessing steps (see more details later in the real data analysis section), we obtain a 
time series of fMRI BOLD signal for each brain region of interest (ROI).  
We then select a set of ROIs and consider the corresponding vector-valued or multivariate time series. 
A sliding window method is used to temporally group 
elements of this multivariate time series into overlapping blocks. 
Within each block we estimate a covariance matrix, and thus obtain a time series of covariance matrices or a SPDM trajectory
for each fMRI recording. This data is then ready for statistical analysis and classification. We describe the main 
pieces of this analysis next.

\subsection{Covariance Estimation}
The first step is to estimate a correlation or a covariance matrix that represents the short-term FC between distinct ROIs in the brain. More specifically, we pick $n$ ROIs and then represent FC 
over a small time-interval using an $n \times n$ covariance matrix of the 
corresponding BOLD signals over that interval. 
In this paper we use the approach from \cite{LEDOIT2004365} to estimate the covariance matrix. 
The estimated $\Sigma$ is an optimal convex linear combination of the sample covariance matrix $S$ and the identity matrix $I$, i.e., $\Sigma=\rho_1I+\rho_2S$, where the optimal weights $\rho_1$ and $\rho_2$ are estimated from the data. The optimality is 
defined with respect to a quadratic loss function, asymptotically as the number of observations and the number of variables go to infinity together. Extensive Monte Carlo simulations confirm that the asymptotic results tend to hold well even in finite sample situations. 
Please refer to \cite{LEDOIT2004365} for more details. 
Now, after estimating a covariance matrix for each time-interval segmented by a sliding window, we obtain a covariance trajectory for each multivariate time series.

\subsection{Riemannian Structure on Symmetric Positive-Definite Matrices (SPDMs)}
In order to quantify differences in FCs, represented by covariance matrices, 
we need a metric structure on the manifold of covariance matrices or SPDMs.  While there are several Riemannian structures in the literature  \cite{Pennec:2006,su-etal-JIVC:2012,zhang-etal:2018}, we use the one introduced in \cite{zhang-etal:2018}, since it has the advantage of having close forms for many necessary operations we need on the SPDM manifold, e.g., geodesic distance, parallel transport, exponential map, inverse exponential map. Zhang et al. \cite{zhang-etal:2018} also have demonstrated that this metric is superior over other metrics such as the log-Euclidean one \cite{Pennec:2006} in analyzing dynamic FCs.   

Let ${\bm {\tilde{ \mathcal{P}}_n}}$ be the space of $n \times n$ SPDMs, and let ${\bm {{\mathcal {P}}_n}}$ be its subset of matrices with determinant one. In our approach, we impose 
separate distances on the determinant one matrices and the determinants themselves.
For any square matrix $G$ with unit determinant, i.e. $G \in GL(n)$, 
we can write it as a product of two square matrices $G = P S$,  
where $P \in {\bm {{\mathcal {P}}}}$ and $S \in SO(n)$  ($SO(n)$ is the set of all $n\times n$ rotation matrices). This is called the 
{\it polar decomposition}. It motivates us to analyze $P$ by representing it as a $G$ after removing $S$. More formally, we identify ${\bm {{\mathcal {P}}}}$ with the 
quotient space $SL(n)/SO(n)$. This identification 
is based on the map $\pi: SL(n)/SO(n) \to {\bm {{\mathcal {P}}}}$, given by 
$\pi([G]) = \sqrt{\tilde{G}\tilde{G}^t}$, for any $\tilde{G} \in [G]$, where the square-root is the symmetric, positive-definite square-root of a symmetric matrix.   The notation $[G]$ stands for all possible rotations of the matrix $G$, given by 
$[G] = \{ GS| S \in SO(n)\}$. 
The inverse map of 
$\pi$ is given by: $\pi^{-1}({{P}}) = [{{P}}] \equiv \{{{P}} S|S \in SO(n)\}$.  This establishes a one-to-one correspondence between the quotient space $SL(n)/SO(n)$ and ${\bm {{\mathcal {P}}}}$. 
Skipping further details, this process leads to the following geodesic distance between points in ${\bm {{\mathcal {P}}}}$. 
For any ${{P}_1}, {{P}_2} \in {\bm {{\mathcal {P}}}}$:
\begin{equation}
\label{eqn:SPDM}
d_{\bm {{\mathcal {P}}}}({{P}_1}, {{P}_2}) =  \|A_{12}\|_F.
\end{equation}
where $A_{12} = \log({ P_{12}})$, 
${ P_{12}} =\sqrt{{{P}_1}^{-1} {{P}_2}^{2} {{P}_1}^{-1}}\ $, 
and $\| \cdot \|_F$ denotes the {\it Frobenious norm} of a matrix. 
Since for any ${\tilde{P}}\in {\bm {\tilde{\mathcal{P}}}}$ we have $\det({\tilde{P}}) > 0$, we can express ${\tilde{P}}$ as a pair $({{P}}, \frac{1}{n}\log(\det({ \tilde{P}})))$ 
with ${{P}} =\frac{{\tilde{P}}}{\det({\tilde{P}})^{1/n}}\ \in {\bm {{\mathcal {P}}}}$. Thus, ${\bm {\tilde{\mathcal {P}}}}$ is identified with the product space of 
${\bm {{\mathcal {P}}}}\times \real_+$ and we take a weighted combination 
of distances on these two components to reach a metric on ${\bm {\tilde{\mathcal {P}}}}$:
\begin{equation}
d_{\bm {\tilde{\mathcal{P}}}}(I, {\tilde{P}})^2 = d_{\bm {{\mathcal {P}}}}(I, { {P}})^2 +\frac{1}{n}\ (\log(\det({\tilde{P}})))^2.
\label{eqn:distI}
\end{equation}	
For any two arbitrary covariances ${\tilde{P_1}}$ and ${\tilde{P_2}}$, let ${\tilde{P}}_{12} = { \tilde{P_{1}}^{-1}}{\tilde{P_2}}S_{12}$ for some optimal $S_{12} \in SO(n)$ 
(using Procrustes alignment). Also, note that for 
${\tilde{P}}_{12} \in {\bm {\tilde{\mathcal{P}}}}$, we have $\det({\tilde{P}}_{12}) = \det(\tilde{{ P_2}})/ \det(\tilde{{ P_{1}}})$. 
Therefore, the resulting squared geodesic distance between ${\tilde{P_1}}$ and ${\tilde{P_{2}}}$ is:  
\begin{equation} 
\label{eqn:distSPDM}
d_{\bm {\tilde{\mathcal{P}}}}({ \tilde{P_1}}, 
{ \tilde{P_2}})^2 = d_{\bm {{\mathcal{P}}}}(I, {{P}_{12}})^2 +\frac{1}{n}\ (\log(\det({ \tilde{P_2}}))-\log(\det(\tilde{ P_1})))^2.
\end{equation}
Under this framework, we also can get closed-form solutions for the exponential map, inverse exponential map, parallel transport and Riemannian curvature tensor, which are often useful. We refer the reader to \cite{su-etal-JIVC:2012} for more details. 

\begin{figure}
	\centering  
	\includegraphics[width=3in]{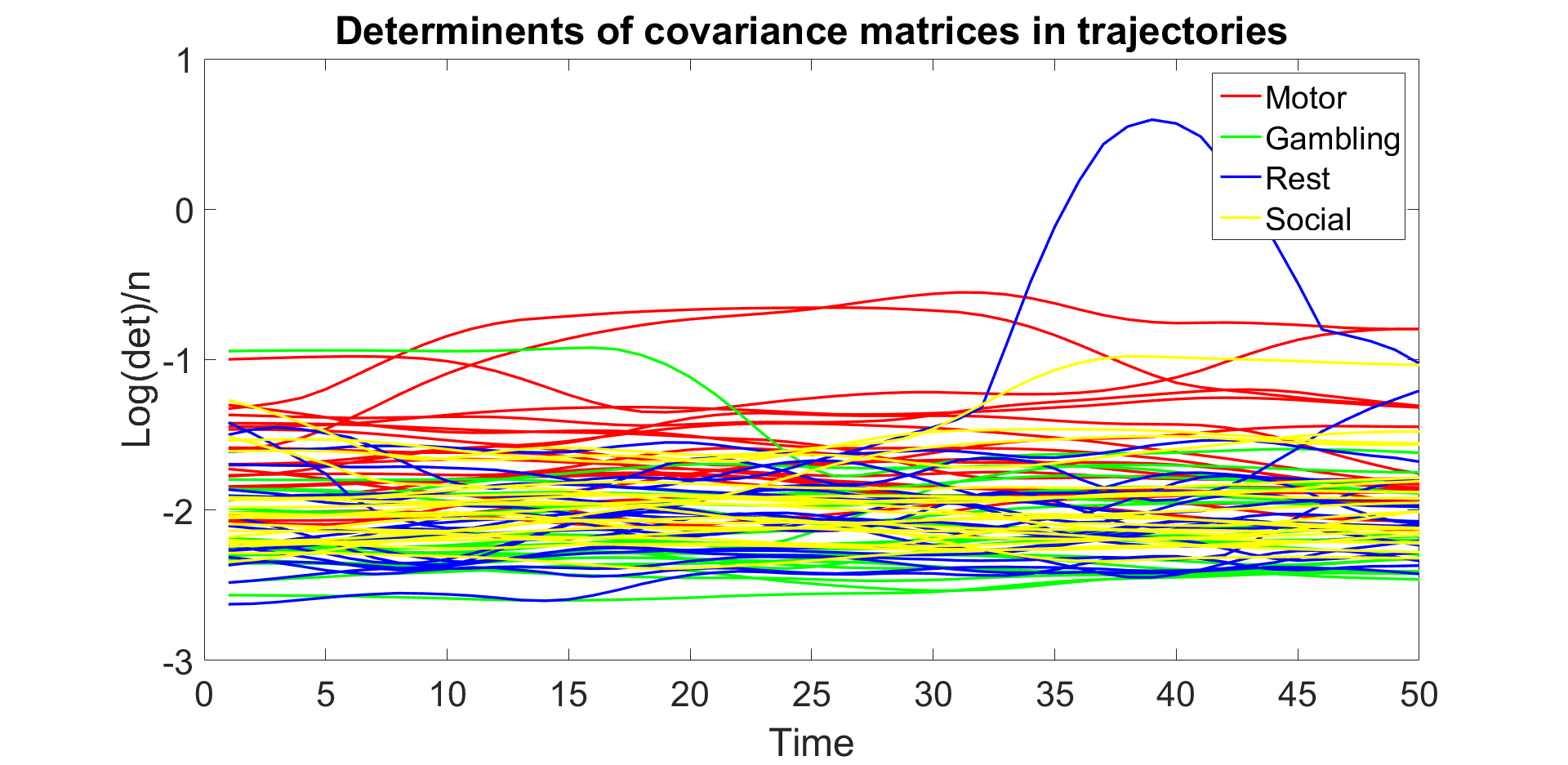}  
	\caption{Determinants of covariance matrices in 80 trajectories from 4 different task activities in the HCP data ($n=4$).} 
	\label{fig:det}
\end{figure}

Note that in Eqn~(\ref{eqn:distSPDM}), distance between covariance matrices is made up of two 
components --  the determinant term and the unit-determinant symmetric matrix term. 
We can choose arbitrary relative weights on these terms to combine the two components. 
While, in some simple cases, it has been shown that one can obtain 
decent classification performance using only the determinant term,  in general 
the second component provides important critical information about actual difference between 
covariance trajectories. 
As an example, Fig.~\ref{fig:det} shows evolutions of  $\log(\det(\alpha(t)))/n$ versus $t$, where 
$\alpha$ denotes a covariance trajectory, for different experimental conditions. It can be seen from 
this plot that the determinant information is not sufficient to classify tasks and experimental conditions. 
Therefore, in the following sections, we will mainly focus on studying the contributions from the unit-determinant symmetric matrix term (first term on the right side of Eqn~(\ref{eqn:distSPDM})). 

\subsection{Riemannian Metric for FC Trajectories} 
To compare FC trajectories from different observations, 
we need a method for calculating distances between such trajectories.
While the previously defined metric on SPDMs provides a way forward, an important issue remains. 
The time rate of executions of different tasks across subjects and experiments may not be
tightly synchronized. Even if the cues are offered at exact times, the reaction times 
of different subjects could be different, even the same subjects can differ in response 
times across experiments. One needs to take care of this execution rate variability in order
to better understand pure FC variation. Mathematically, this variability is represented as follows. 

Let $\alpha$ denote a smooth trajectory on the Riemannian manifold of SPDMs $\bm {\mathcal{P}}$, where $\bm {\mathcal{P}}$ is endowed with the Riemannian distance in (\ref{eqn:distI}). 
Let ${\mathcal{M}}$ denote the set of all such trajectories: 
${\mathcal{M}} = \{\alpha : [0, 1] \to {\bm {\mathcal{P}}}| \alpha $ is smooth$\}$. 
Also, define $\Gamma$ to be the set of all diffeomorphisms of $[0, 1]: \Gamma = \left\{ \gamma : [0, 1] \to [0, 1]| \gamma(0) = 0, \gamma(1) = 1 \right\} $, 
$\gamma$ is a diffeomorphism with the end points preserved. 
Elements of $\Gamma$ represent different execution rates across different experimental recordings. 
If $\alpha$ is a trajectory on $\bm {\mathcal{P}}$, then $\alpha \circ \gamma$ is a trajectory that follows the same sequence of points as $\alpha$ but at the evolution rate governed by $\gamma$. That is, $\alpha$ and $\alpha \circ \gamma$ differ
only in the execution rates, or response times, the actual FC responses are identical. 
It is important to note that $\Gamma$ forms a group under the composition operation.

Let $\alpha_1$ and $\alpha_2$ be two smooth trajectories in ${\mathcal{M}}$, 
a simple way to establish a metric between them is to use 
$\int_{0}^{1}d(\alpha_1(t),\alpha_2(t)) dt$. 
Although it represents a natural solution to quantify differences 
between $\alpha_1$ and $\alpha_2$, it suffers from the problem that
$d(\alpha_1, \alpha_2) \neq d(\alpha_1 \circ \gamma_1, \alpha_2 \circ \gamma_2)$, 
when the activities are observed under different execution rates denoted by $\gamma_1$ and $\gamma_2$.  
This implies that even the same FC trajectories will have non-zero distances 
between them if they are performed at different execution rates. 
The execution rate $\gamma$ is often considered as a nuisance variable 
if we only consider the value or the shape of the trajectories.
We will later demonstrate that removing the effects of $\gamma$ will sometime 
facilitate better classification of FCs in certain scenarios.  
To make comparison of trajectories invariant to their
execution rates, we use the Riemmanian metric as described in  \cite{zhang-etal:2018}. This construction
is described next.

{\bf Definition 1} 
Let $\alpha : [0, 1] \to {\bm  {\mathcal{P}}}$ be a smooth trajectory 
with a starting point $\alpha(0) = P \in  {\bm  {\mathcal{P}}}$.  
Define the transported square-root vector field
(TSRVF)  of $\alpha$ to be a scaled parallel-transport of the vector field 
$\dot{\alpha}$ along $\alpha$ to the starting point $P$ according to: 
for each $t \in [0, 1]$,
$q(t) = \frac{\dot{\alpha}(t)_{\alpha(t) \to P}}{\sqrt{|\dot{\alpha}(t)|}} \in \mathcal{T}_P (\bm  {\mathcal{P}})$, 
where $| \cdot |$ denotes the norm that is defined through the Riemannian metric on $\bm  {\mathcal{P}}$.
$V_{\alpha(t) \to \alpha(0)}$ denote the parallel transport of $V$ from $\alpha(t)$ to $\alpha(0)$ along $\alpha$.
A covariance trajectory $\alpha$  is then represented by the pair $(P, q(t))$ in this analysis. 

For any point $P \in \bm  {\mathcal{P}}$, denote the set of all square-integrable
curves in $\mathcal{T}_P (\bm {\mathcal{P}})$ as $\mathbb{C}_P \equiv L^2
([0, 1], \mathcal{T}_P (\bm {\mathcal{P}}))$. The space of all smooth SPDM trajectories, then, becomes an infinite-dimensional vector bundle $\mathbb{C} = \amalg_{P \in {\bm {\mathcal{P}}}} \mathbb{C}_P,$ which is the indexed union of $\mathbb{C}_P$ for every $\mathcal{T}_p (\bm {\mathcal{P}})$. 

For given two points $(P_1,q_1)$ and $(P_2,q_2)$ on $\cC$, we want to find the geodesic path 
connecting them. Let $( x(s),$ $v(s,\cdot) ), s\in[0,1]$ be a path with 
$\left( x(0),v(0,\cdot) \right) = (P_1, q_1)$ and $\left( x(1),v(1,\cdot) \right) = (P_2, q_2)$. To become a geodesic, $( x(s),$ $v(s,\cdot) )$ needs to satisfy certain conditions according to \cite{zhang-etal:2018,le2018discrete} and these conditions help 
develop a geodesic shooting method for finding geodesic between two points on $\cC$.
Skipping details, we provide the final expressions for geodesic distances. 

{\bf Defnition 2}
Given two trajectories $\alpha_1$, $\alpha_2$ and their representations $(P_1, q_1),(P_2, q_2) \in \mathbb{C}$, and
let $(x(s), v(s)) \in \mathbb{C}, s \in [0, 1]$ be  the geodesic between $(P_1, q_1)$ and $(P_2, q_2)$ on $\mathbb{C}$, the geodesic distance is given as:
\begin{equation}
\label{eqn:disttraj}
d_c((P_1, q_1),(P_2, q_2)) =\sqrt{{l_x}^2+\int_{0}^{1}|{q_1}^{\parallel}(t)-q_2(t))|^2 dt}.
\end{equation}
This distance has two components: (1) the length between the starting points on $\bm {\mathcal{P}}, l_x = \int_{0}^{1}|\dot{x}(s)| ds$; and (2) the standard $L
^2$ norm on $\mathbb{C}_{p_2}$ between the TSRVFs of the two trajectories,
where ${q_1}^{\parallel}$ represents the parallel transport of $q_1 \in \mathbb{C}_{p_1}$ along $x$ to $\mathbb{C}_{p_2}$. In the following context, we will also denote $d_c((P_1, q_1),(P_2, q_2))$ as $d_c(\alpha_1, \alpha_2)$ for simplicity. 

The main motivation for us to use the TSRVF representation, to represent a SPDM trajectory, is that we have the following important property held: $d_c(\alpha_1, \alpha_2) = d_c(\alpha_1\circ \gamma, \alpha_2 \circ \gamma)$. This key property allows us to study dynamic FCs in a manner that is invariant to their evolution rate. That is we can compare two equivalence classes or orbits $[\alpha_1]$ and $[\alpha_2]$, where $[\alpha_i] = \{\alpha_i \circ \gamma| \gamma \in \Gamma\} $.  

{\bf Definition 3}
Define the geodesic distance $d_q$ on $\mathbb{C}/\Gamma$ as the shortest distance between two orbits in $\mathbb{C}$, 
\begin{equation}
\label{eqn:distalign}
d_q([\alpha_1],[\alpha_2]) = \inf_{\gamma \in \Gamma} 
d_c((P_1, q_1),(P_2, (q_2 \circ \gamma)\sqrt{\dot{\gamma}})).
\end{equation}
It has been shown in \cite{zhang-etal:2018} that this is a proper metric in the space $\mathbb{C} / \Gamma$. To solve the above equation, we take the following optimization strategy
\begin{equation}
\label{eqn:geo}
min_{(x,v),\gamma}({l_x}^2+\int_{0}^{1}|{q_{1,x}}^{\parallel}-(q_2 \circ \gamma)\sqrt{\dot{\gamma}}|^2 dt)
\end{equation}
This optimization is very expensive since we need to iteratively search for the optimal path $(x,v)$ on $\cC$ and the re-parameterization $\gamma$. 
Instead, we use a fast approximation here. 
We first find the baseline $x$ connecting two trajectories using geodesic between 
$\alpha_1(0)$ and $\alpha_2(0)$ on $\bm {\mathcal{P}}$ and then align their TSRVFs accordingly using the Dynamic Programming Algorithm. This substantially speeds up the 
solution albeit at the cost of potentially differing from the optimal solution stated under the theoretical formulation.

\subsection{Dimension Reduction for SPDMs}
As mentioned above, the number of ROIs in brain connectivity studies 
can become very large \cite{ARSLAN20185}, resulting in FC trajectories with
large size SPDMs. The problem of comparing trajectories can become 
computationally expensive and, therefore, we seek a method for 
the data reduction, while preserving the symmetric, positive-definite nature of covariance matrices. Our main idea is to find a linear projection that maps high-dimensional SPDMs to low-dimensional 
SPDMs in a principled, near-optimal manner. 
In addition to providing computational simplification, the low-dimensional SPDMs also facilitate analyses in the following ways: 
\begin{enumerate}
	\item  Such a projection 
	can bring FC trajectories associated with different number of ROIs to the same smaller dimension, and make comparisons between them possible.
	\item  In the case that not all ROIs carry the same amount of information,  dimension reduction can help filter out some 
	extraneous components.
\end{enumerate}

The problem of dimension reduction of 
SPDMs has been studied and used in a variety of computer vision and pattern recognition problems, 
see e.g., \cite{Harandi2017DimensionalityRO, DBLP:journals/corr/LiL17}. 
In this paper, we develop a novel dimension-reduction technique adapted to the Riemannian metric presented in Section II B. Thus, the reduced SPDMs are especially suitable for analyzing under the proposed Riemannian framework. 

Given a set of $n \times n$ unit-determinant SPDMs  $\{ {P}_i \}$, where $n$ is a large integer, 
our goal is to find orthogonal 
matrix ${B} \in \real^{n \times d}$, where $d<< n$ and ${B^T}{B} = I_d$, 
to project ${P}_i$ to $Q_i$ in $\real^{d \times d}$ according to  $ Q_i=B^T{{P}_i}B$. The space of such 
orthogonal matrices is called a {\it Stiefel manifold}, denoted by ${\cal S}_{n,d}$. 
The next question is: What should be the optimality criterion for defining $B$?
A simple yet important idea is that pairwise distances 
between given SPDMs should be preserved as much as possible after the projection. 
That is, find ${ B} \in {\cal S}_{n,d}$ such that $d_{ \bm {{\P}_d}} (Q_i,Q_j) \approx 
d_{ \bm {{\P}}_n}({P}_i,{P}_j)$ for all $i, j$ in 
the training set. This criterion can be formulated as:
$$\argmin_{B \in {\cal S}_{n,d}} \sum_{i,j}
\left( d_{\bm {{\P}_n}}({P}_i,{P}_j) - d_{\bm {{\P}_d}}(Q_i,Q_j)\right)^2
\ .
$$ 
A direct optimization of this quantity over ${B} \in {\cal S}_{n,d}$
is complicated due to the complexity of
the chosen Riemannian metric. Instead,  
we develop an approximation where the comparison of distances is replaced by the 
comparison of relevant matrices directly.

In the original space, the distance between matrices $P_i$ and $P_j$ is governed by the matrix 
$P_{ij} = P_i^{-1} P_j^2 P_i^{-1}$ (see Eqn~(\ref{eqn:SPDM})). Similarly, distance in the 
smaller space is determined by the matrix $Q_{ij} = Q_i^{-1} Q_j^2 Q_i^{-1}$. In order to compare
these matrices, we need to bring them to the same space. 
Let $\hat{P}_i$ denotes the reconstruction of $P_i$ from 
its projection $Q_i$, i.e.  $\hat{P}_i = BQ_i B^T \in \real^{n \times n}$. 
Our goal is to find $B \in {\cal S}_{n,d}$ that minimizes the quantity: 
\begin{equation}
\argmin_{B \in {\cal S}_{n,d}} \sum_{i,j} \| P_{ij}  -  \hat{P}_i^{-1} \hat{P}_j^2 \hat{P}_i^{-1}\|^2 .
\label{eqn:opt0}
\end{equation}
However, this specification requires the following proviso. 
Since $\hat{P}_i$ is rank $d$, it is not invertible, and one needs to use its pseudoinverse
instead. Let $\hat{P}_i^{-} = B Q_i^{-1} B^T$ denote the pseudoinverse of $\hat{P}_i$. 
(In the appendix we establish this expression as a pseudoinverse.)
Then, we have the following result.
\begin{lemma}
	Under the conditions specified above, we have, for all $i$, $j$, 
	$$
	\| P_{ij} - \hat{P}_i^{-} \hat{P}_j^2 \hat{P}_i^{-}\|^2
	= \| P_{ij} -   BQ_{ij}{B}^T \|^2 .
	$$
	\label{lemma:lm1}
\end{lemma}
To prove this, one only needs to show that $\hat{P}_i^{-} \hat{P}_j^2 \hat{P}_i^{-} = BQ_{ij}{B}^T$, which is proven in the appendix. This provides another interpretation of the objective function. 

\begin{lemma}\label{lemma:lm2}
	The optimization of quantity in Lemma~\ref{lemma:lm1} can be rephrased as follows.
	\begin{eqnarray*}
		&&	\argmin_{B \in {\cal S}_{n,d}} \sum_{i,j=1}^N \| P_{ij} - BQ_{ij}{B}^T \|^2\\
		&&= \argmax_{B \in {\cal S}_{n,d}}  \left( \sum_{i,j=1}^N tr({B^T}{ P}_{ij} {B} {B^T} {P}_{ij} B) \right)\ .
	\end{eqnarray*}
	
\end{lemma}
The proof of Lemma~\ref{lemma:lm2} is given in the appendix. 
We solve this optimization problem on the Stiefel manifold using the 
Matlab toolbox  {\it Manopt} \citep{Boumal:2014}.

\section{Experiments}
In this section, we provide a number of simulations and real data experiments to illustrate the proposed methodology. 
The first simulated example illustrates the proposed 
dimension-reduction technique. 
The second simulation studies the combined effects of dimension reduction and 
temporal alignment. In the real data experiment section, 
we present a few classification experiments based on different ROIs, activities, 
and classifiers with and without dimension reduction and alignment.

\subsection{Simulation Studies on Dimension Reduction and Alignment}

\begin{figure*} 
	\centering
	\begin{tabular}{ccccc}
		\includegraphics[width=1.4in]{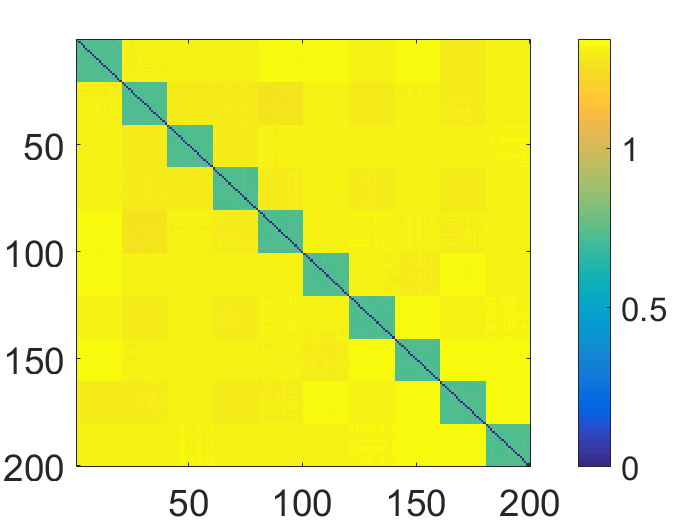}&
		\includegraphics[width=1.4in]{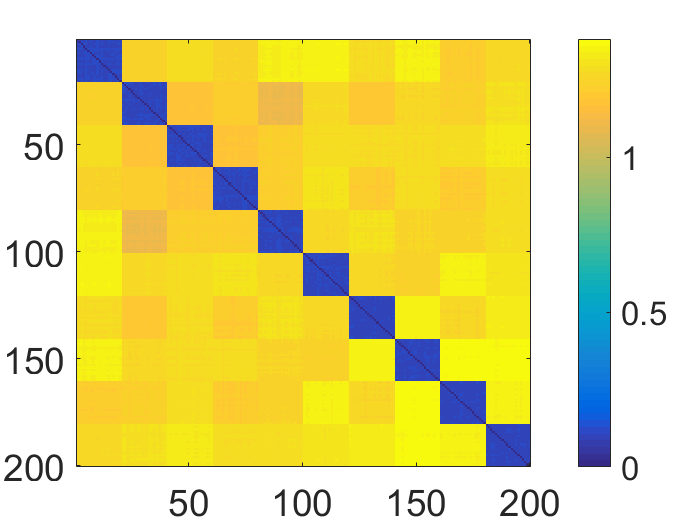}&
		\includegraphics[width=1.4in]{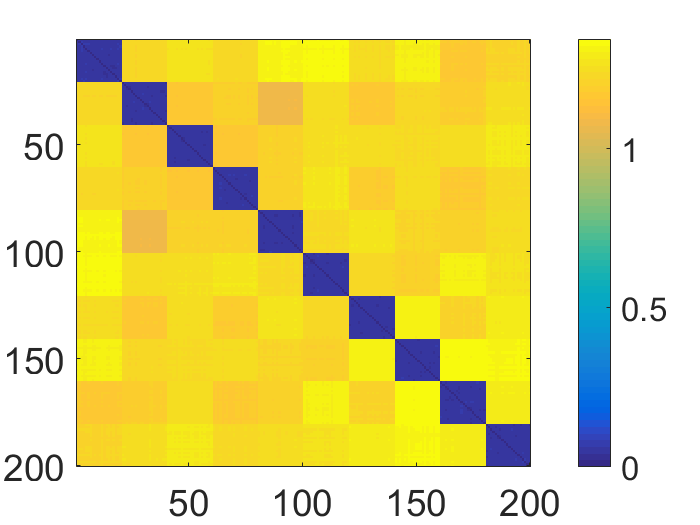}&
		\includegraphics[width=1.4in]{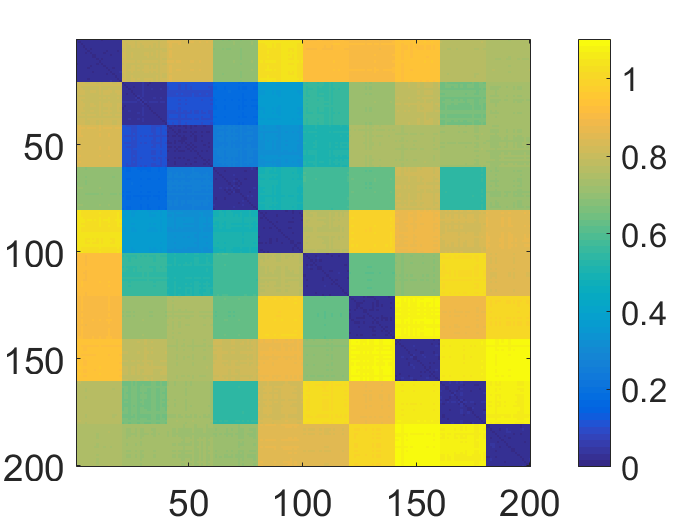}&
		\includegraphics[width=1.4in]{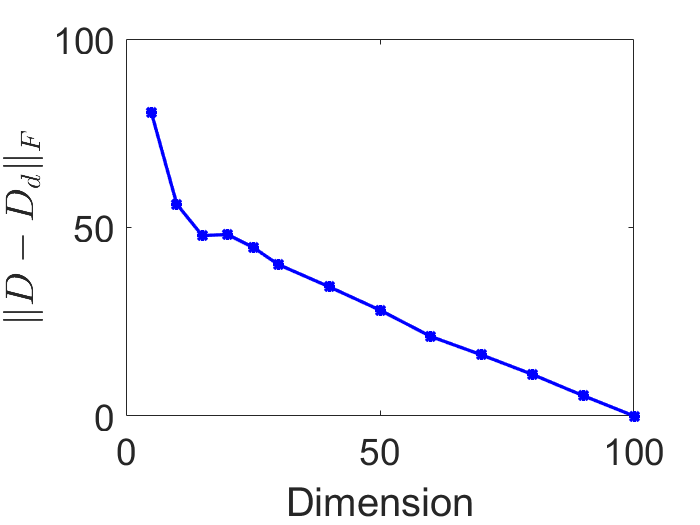}	\\
		(a) & (b) & (c) & (d) & (e)\\
	\end{tabular}
	\caption{Pairwise distances between 200 covariance matrices before and after dimension reduction. (a) shows the distance matrix before dimension reduction, and (b), (c), (d) show the distance matrices after reducing the dimension to d=20, 10 and 5, respectively. (e) shows $ \|D - D_d \|_F $ {\it vs} dimension.} 
	\label{fig:simulation1}
\end{figure*}

\begin{figure*}[h]
	\centering  
	\includegraphics[width=5in]{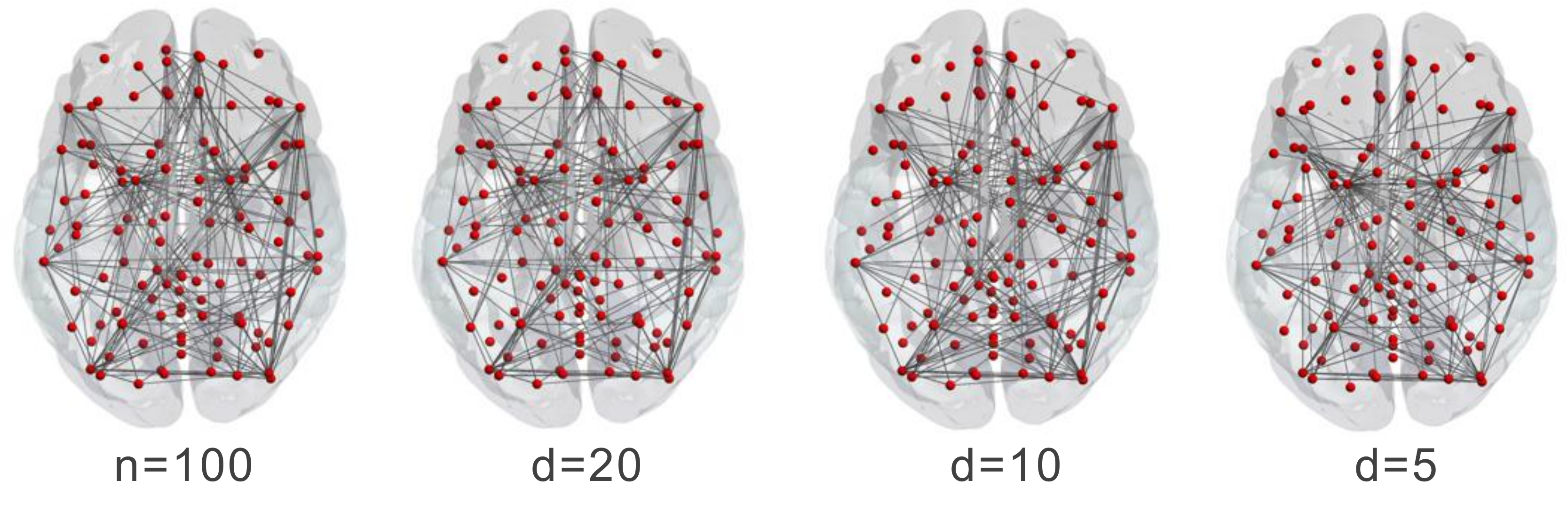}  
	\caption{Visualization of connectivities using precision matrices estimated by reconstructed SPDMs from different dimensions.} 
	\label{fig:recon_display}
\end{figure*}

\noindent {\bf Simulation Experiment 1}: We first randomly generate $k$ sets of SPDMs, each 
set containing $T$ independent $n \times n$ SPDMs. Let ${P_i}^{(j)}$ be the $j$th SPDM in the $i$th set; 
it is generated using ${P_i}^{(j)}=K_i {K_i}^T + nI_n + \varepsilon_{ij}{\varepsilon_{ij}}^T$, 
where $K_i$ and $\varepsilon_{ij}$ are $n \times n$ square matrices generated with 
{\it i.i.d} standard normal entries. In this simulation we have used the values 
$n=100$, $T=20$ and $k=10$. Next, 
we apply our dimension reduction method to reduce the dimension of SPDMs from $n=100$ to 
$d=20, 10$ and $5$. In order to evaluate this reduction, 
we calculate pairwise-distance matrices before and after dimension reduction, and display them in 
Fig.~\ref{fig:simulation1}.  The results show that although the dimension 
of SPDM changes dramatically, 
the patterns of distances within and across sets are preserved well for 
$d = 20$ and $d=10$, with a slight deterioration 
in the case of $d = 5$. To quantify changes in distance matrices, we let $D$ be the distance matrix between covariance matrices in the original dimension, $D_d$ be the distance matrix between covariance matrices after reducing dimension to $d$, and display their Frobenius norm $ \|D - D_d \|_F $ {\it vs} $d$ in Fig.~\ref{fig:simulation1} (e). We also utilize some real data associated with social task along with AAL \cite{AAL} template to display functional connectivities using reconstructed SPDMs in different dimensions, with BrainNet Viewer \cite{10.1371/journal.pone.0068910}.  We threshold precision matrices associated with these SPDMs to represent edges in graphs, as shown in Fig.~\ref{fig:recon_display}. Similarities between these graphs indicate the success of data compression with our dimension reduction method. 
\\

\noindent {\bf Simulation Experiment 2}: 
Next, we study the combined effects of dimension reduction and 
temporal alignment. 
We simulate covariance trajectories in the space of large SPDMs and time warp them to generate new 
trajectories. Then, we perform dimension reduction and solve for temporal registration in the reduced
space, in that order, and compare the warping results to the original time warpings. 

\begin{figure}
	\centering  
	\begin{tabular}{ccc}
		\includegraphics[width=1.15in]{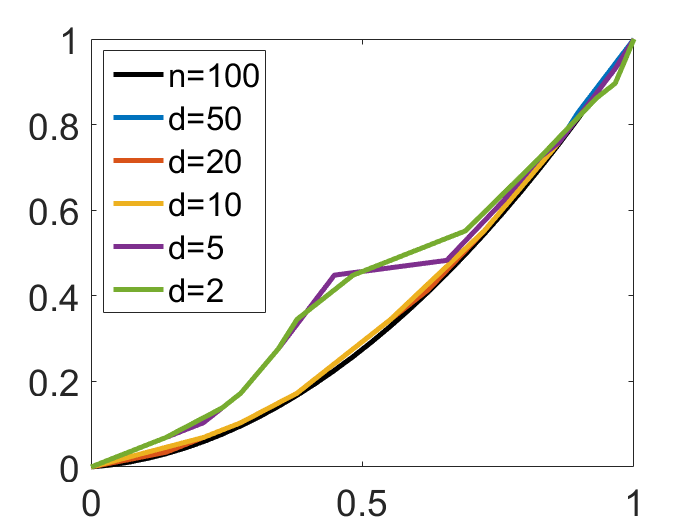}
		&
		\includegraphics[width=1.15in]{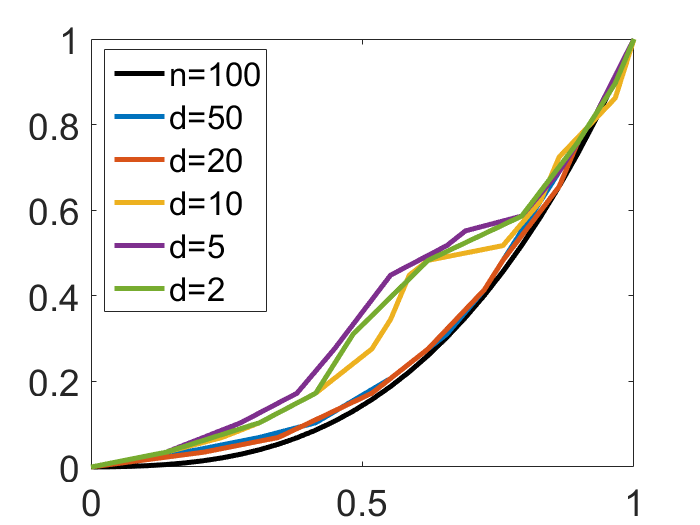}
		&
		\includegraphics[width=1.15in]{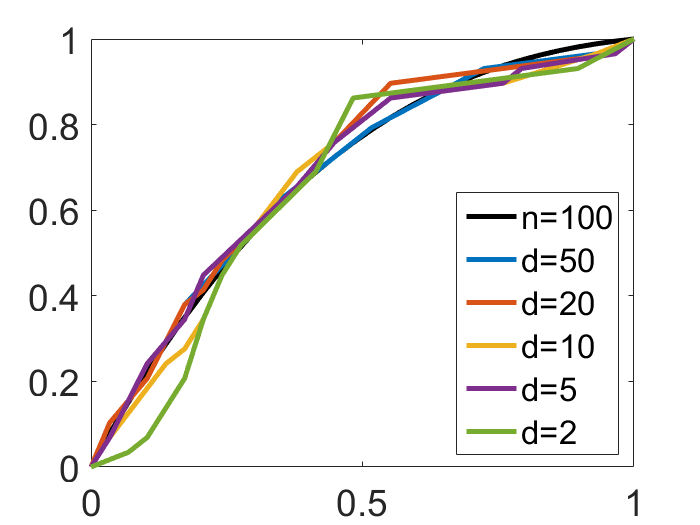}	\\
		(a) & (b) &(c)\\
	\end{tabular}  
	\caption{The recovered warping functions before and after different levels of dimension reduction. x-axis and y-axis represent time before and after warping respectively.} 
	\label{fig:simulation2}
\end{figure}

We start by generating several covariance sequences, labeled 
$\{S_i\}$, each with $20$ time points and the covariance size $100 \times 100$. 
They are generated as follows: 
First we simulate a time series in $\real^{100}$ from multivariate normal distribution 
${\cal N}(0,I_{100})$ for $t = 1,..., 300$, and then use a sliding window with window size of 
$80$ and step size of $10$  to segment the time series into overlapping blocks. 
We then estimate covariance in each block to form a $20$-length sequence of 
$100 \times 100$ covariances. The resulting sequence is resampled at $20$ 
time points after temporal  
smoothing with a Gaussian kernel resulting in $S_i$. Next, we randomly generate warping functions $\gamma_i$, and 
form new sequences $S_i \circ \gamma_i$, denoted by $\tilde{S}_i$.
Finally we implement dimension reduction on $S_i$ and $\tilde{S}_i$, respectively to obtain
trajectories of size $d \times d \times 20$, with $d <<100$, denoted by $S^d_i$ and $\tilde{S}^d_i$. 
Then, we use the optimization stated in Eqn~(\ref{eqn:geo}) to align 
$S^d_i$ to $\tilde{S}^d_i$. In Fig. ~\ref{fig:simulation2}, we show the recovered warping functions
for different cases. Each panel represents a different experiment with different simulated data. 
The ground truth warping function is shown in black color for $n=10$, while the other curves 
represent the recovered warping functions for different values of $d$. 
From these results, we can see that for a moderate dimensional reduction, 
with $d = 50$ or $d = 20$, the alignment results are near perfect. 
However, a large reduction ($d \leq 5$) can result in some loss of information and 
lead to a deterioration in alignment performance.   

\begin{figure}[h]
	\centering  
	\includegraphics[width=2in]{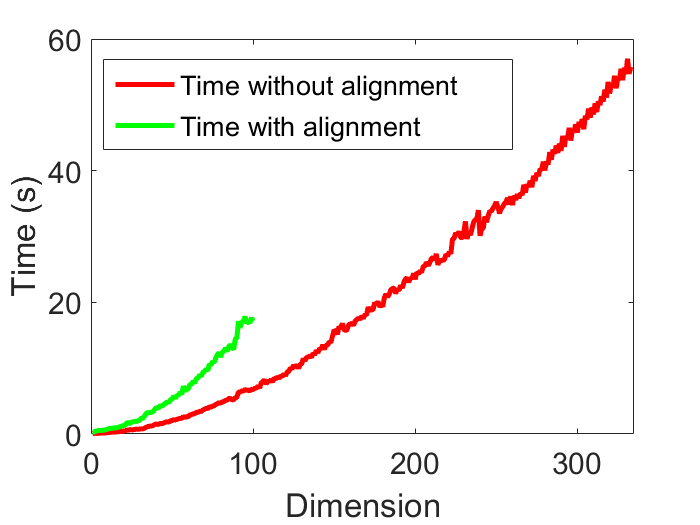}  
	\caption{Time cost for calculating distance between two $333 \times 333 \times 20$ trajectories before and after alignment with different degrees of dimension reduction.} 
	\label{fig:timevsdim}
\end{figure}

The computational cost for analyzing a large set of SPDM trajectories is
a big concern, thus motivating the need for dimension reduction. In Fig.~\ref{fig:timevsdim} we 
plot the computational cost of comparing trajectories, with and without temporal alignments,  
for different sizes of SPDM trajectories. This cost was computed with  
a PC with Intel i7-6700HQ and 8GB RAM. One can see that the cost of performing temporal 
alignment grows rapidly with the matrix size and becomes impractical for $n$ larger 
than $100$.
This is an important consideration and sometime the gains of alignment are nullified by increases in the computational cost.


\subsection{Real Data Experiments}
Next we use HCP dataset to perform quantitative analysis of fMRI responses
from different ROIs under different activities. The majority of the HCP fMRI data were acquired at 3T. A series of 4D fMRI imaging data were acquired for each subject while they were performing different tasks involving different neural systems, e.g. visual, motion and cognition systems. The acquired image is with an isometric spatial resolution of $2$ mm and temporal resolution of $0.7$ s. All fMRI data in HCP are preprocessed by removing spatial distortions, realigning volume to compensate for subject motion, registering the fMRI to the structural MRI, reducing the bias field, normalizing the 4D image to a global mean, masking the data with the final brain mask and aligning the brain to a standard space (MNI space) \cite{GLASSER2013105}. This preprocessed fMRI data are then available for connectivity analysis. 

We begin by segmenting brain into multiple ROIs 
using an existing template \cite{Gordon2016}. With this template, the cortical area is divided into several networks: 
visual, auditory, default, frontoparietal, dorsal attention, cingulo-opercular, ventral attention, and salience. 
The fMRI time series data for each region (totally, 333 regions) are extracted using CONN functional connectivity toolbox \cite{pmid22642651}. Similar to the simulation study, we use a sliding window method to 
reach covariance trajectories representing dynamic FCs.  For each activity we use the first 176 time points in fMRI time series. Despite the incompleteness of data, this setup makes sure the resulting sequences are compared at the same time scale.  In all we extract data from 196 subjects under five different tasks - motor, gambling, social cognition, language, emotion -  and the resting state and use them in our analysis.

In each classification experiment, we calculate pairwise distance 
$d_q$, as defined in Eqn~(\ref{eqn:distalign}),
between individual trajectories, and feed these distances into two types of classifiers.
One is SVM with RBF kernel. The other is a simple deep neuron network (DNN) classifier with four dense layers along with dropout and ReLu activation at each layer, and one last dense layer with softmax activation, using Keras with Tensorflow \citep{tensorflow2015-whitepaper} backend. We find in some cases alignment helps although that, in the more general multiclass classification tasks, using $d_c$ in Eqn~(\ref{eqn:disttraj}) we can still obtain same level of accuracies. Thus, one can choose to use Eqn~(\ref{eqn:disttraj}) in these tasks to save computation time.
The results of individual experiments are presented next.

\subsubsection{Dynamic FCs under different sets of ROIs and a fixed task}
\begin{figure}[tbh]
	\centering  
	\begin{tabular}{ccc}
		\includegraphics[width=1.15in]{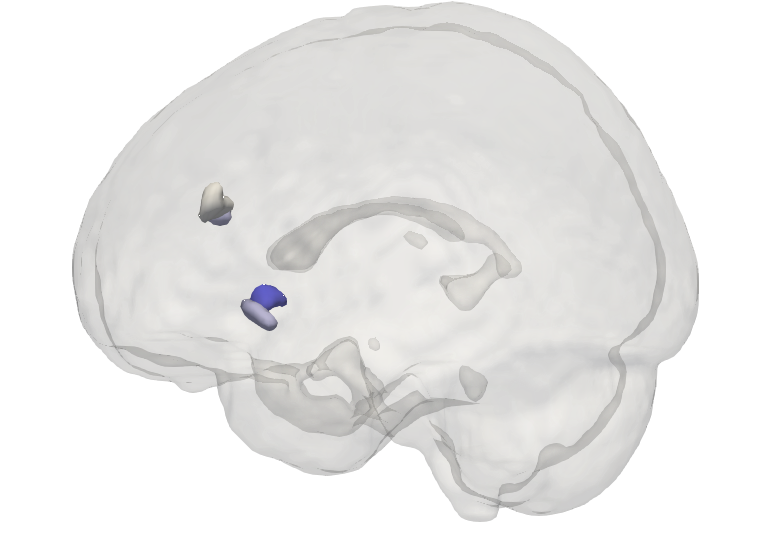}&
		\includegraphics[width=1.15in]{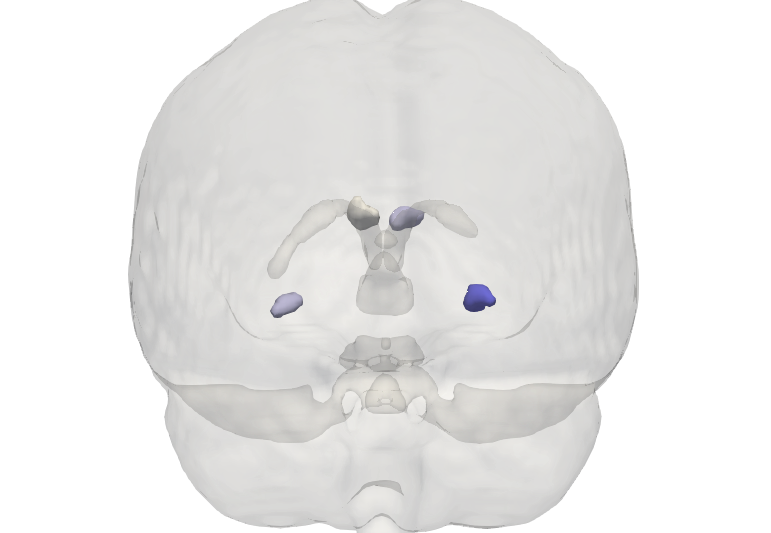}&
		\includegraphics[width=1.15in]{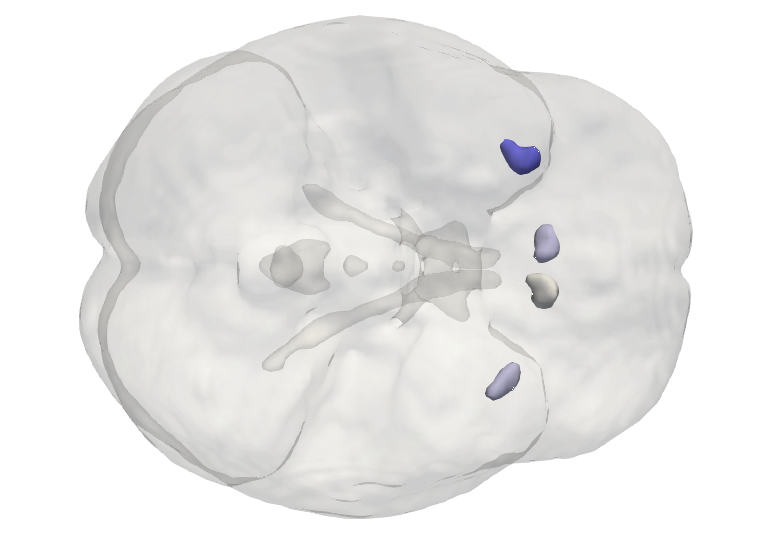}
		\\
		&(Salience)\\
		\includegraphics[width=1.15in]{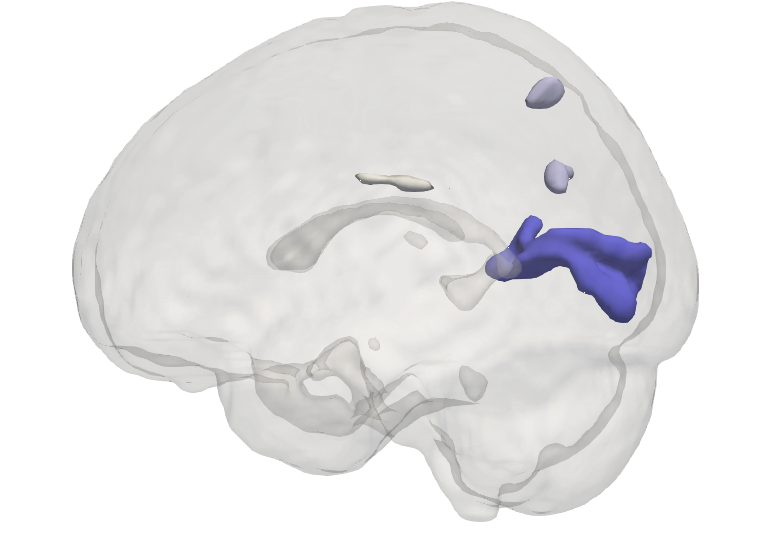}&
		\includegraphics[width=1.15in]{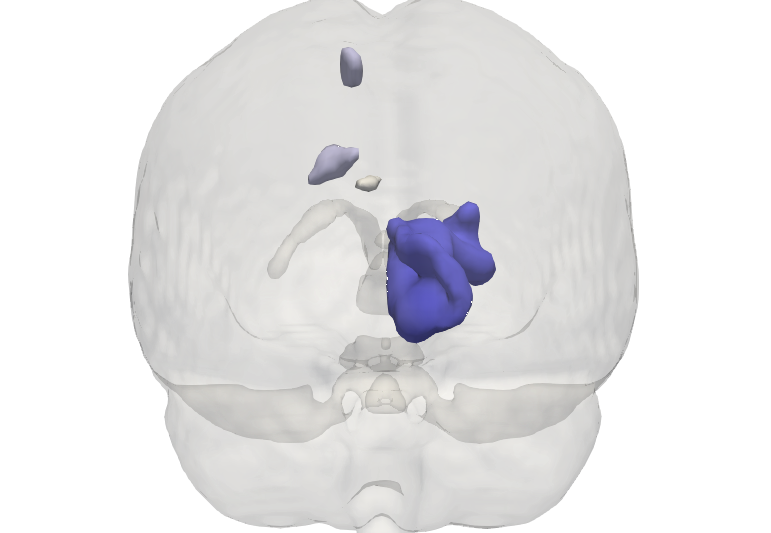}&
		\includegraphics[width=1.15in]{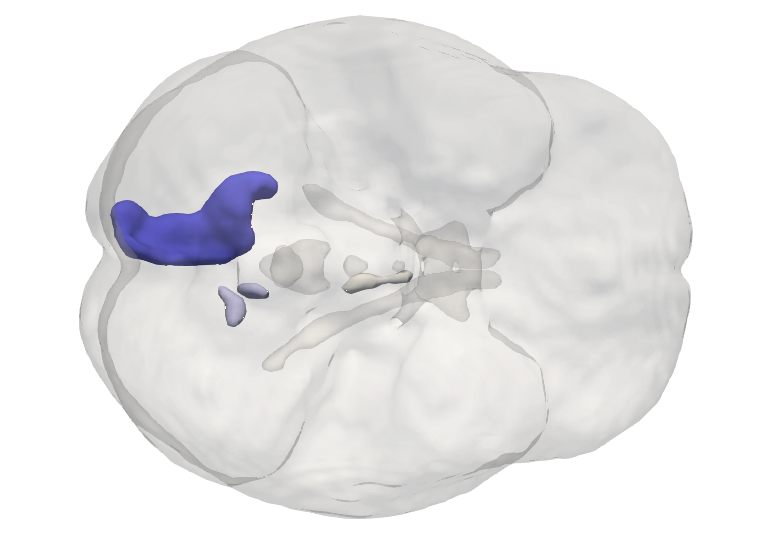}
		\\
		&(CinguloParietal)\\
		\includegraphics[width=1.15in]{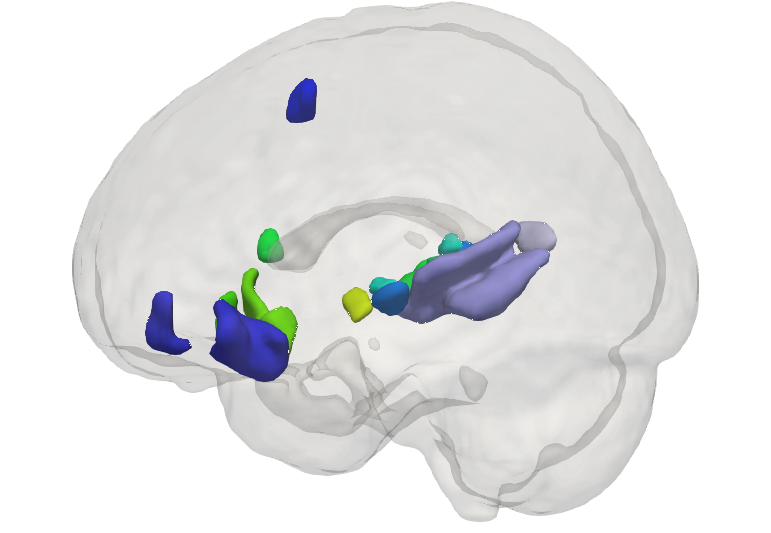}&
		\includegraphics[width=1.15in]{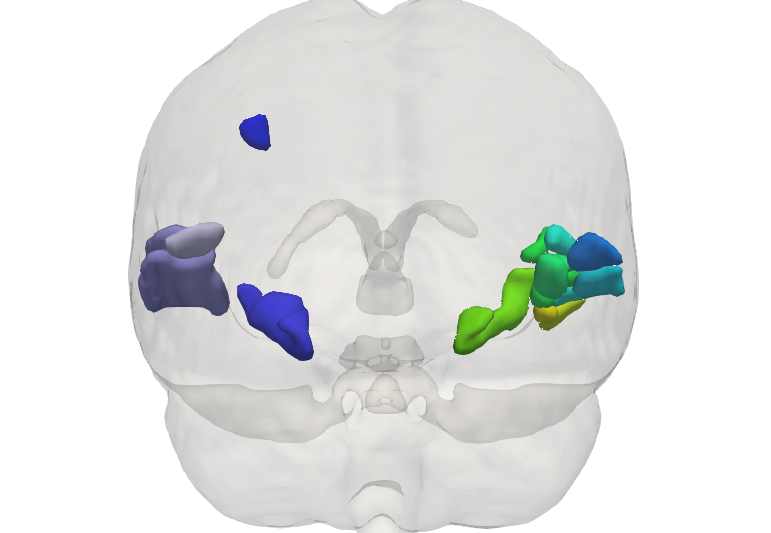}&
		\includegraphics[width=1.15in]{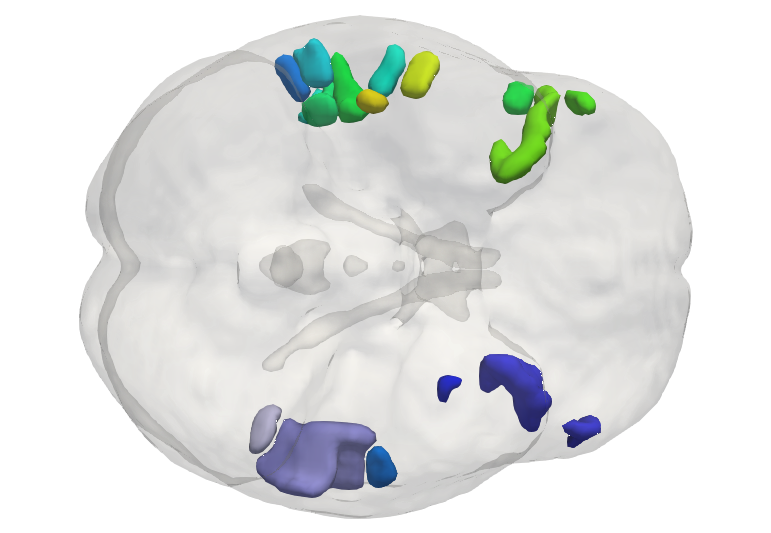}
		\\
		&(VentralAttn)\\
		\includegraphics[width=1.15in]{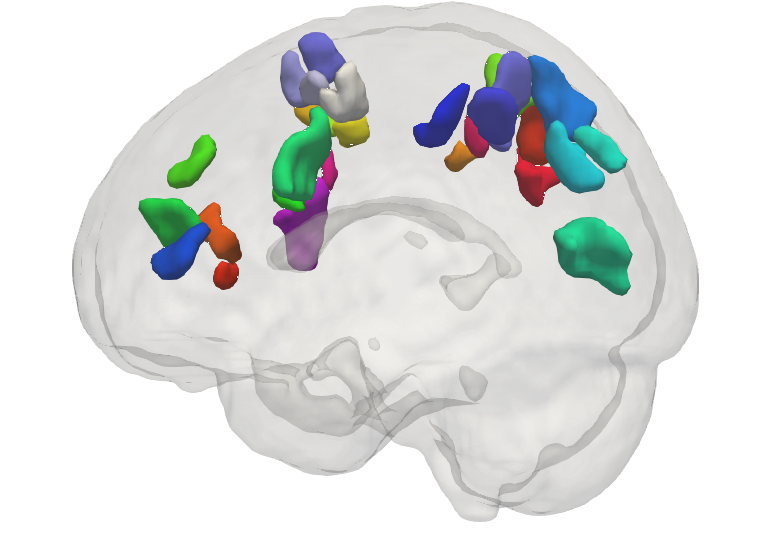}&
		\includegraphics[width=1.15in]{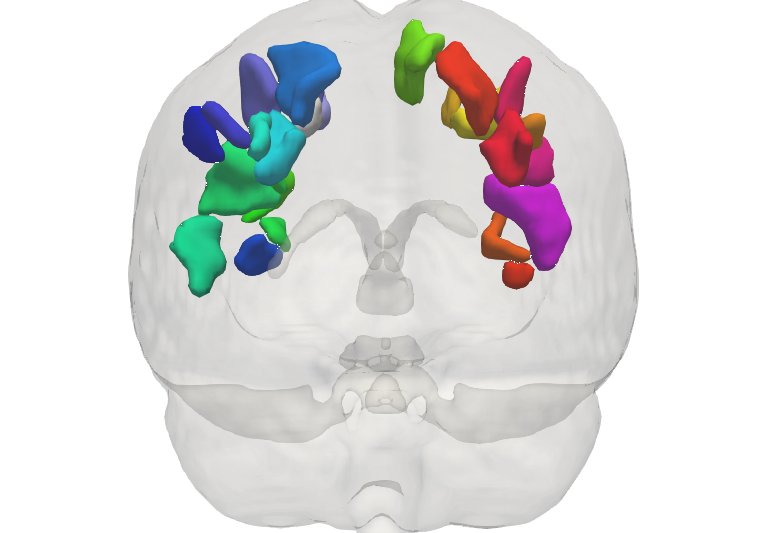}&
		\includegraphics[width=1.15in]{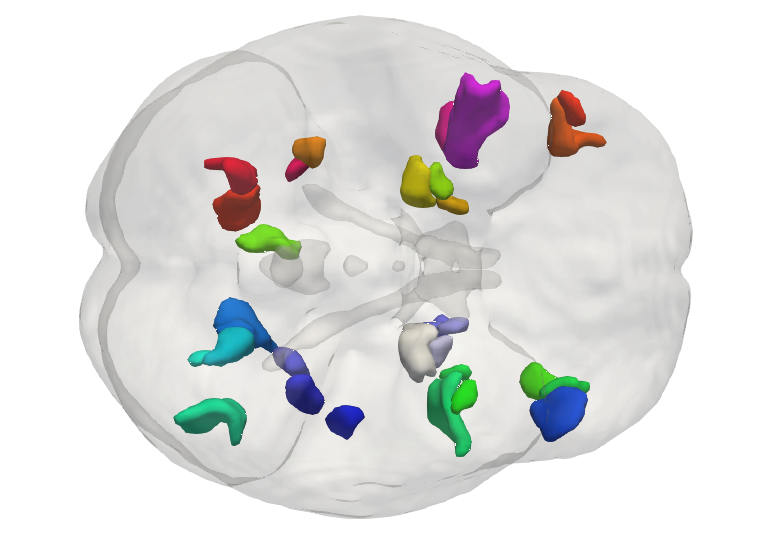}
		\\
		&(DorsalAttn)\\
	\end{tabular}
	\caption{Plots of ROIs in the Salience, CinguloParietal, VentralAttn and DorsalAttn networks in the used template. } \label{fig:plotROIs}
\end{figure}

\begin{figure}
	\centering  
	\begin{tabular}{ccc}
		\includegraphics[width=1.15in]{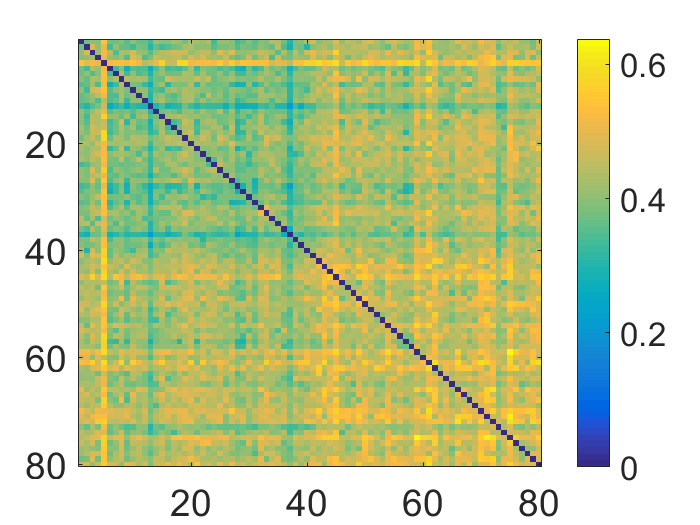}&
		\includegraphics[width=1.15in]{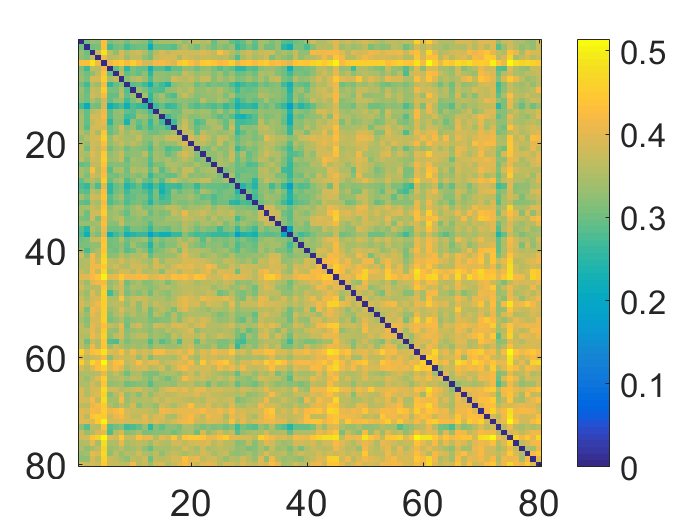}	&
		\includegraphics[width=1.15in]{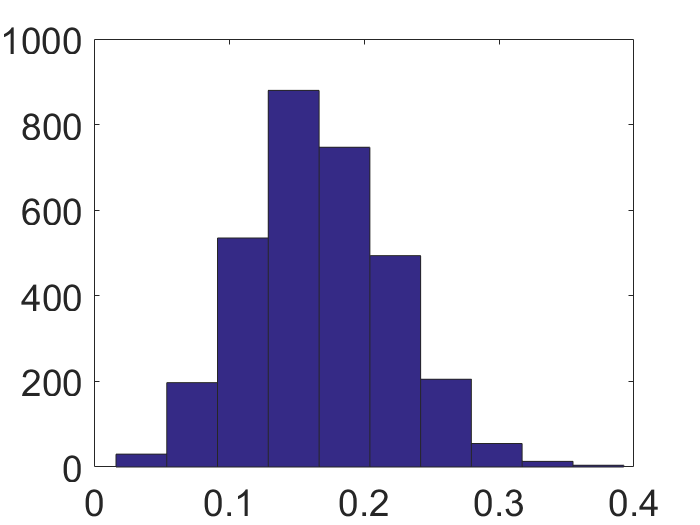}
		\\
		(a) & (b) & (c)\\
	\end{tabular}
	\caption{Pairwise distance matrix between the 80 trajectories from the gambling task. Group 1: first 40 trajectories from 4 ROIs in Salience network involving 4 ROIs; Group 2: second 40 trajectories from 4 ROIs in CinguloParietal network. (a) Pairwise distances before alignment. (b) Distances after alignment (c) Histogram of ($d_c$-$d_q$)/$d_c$.
	} \label{fig:gamblingdist}
\end{figure}

In this experiment, we restrict to the gambling task and use 
different sets of ROIs to study corresponding 
dynamic FCs. The first set of 4 ROIs is from the Salience network and 
the second set of 4 ROIs is from the CinguloParietal network. Fig.~\ref{fig:plotROIs} shows locations of these ROIs in 
human brain.  We extract the fMRI signals from 40 subjects in the HCP data set, and in total we obtain $80$ trajectories.
40 for each set of ROIs. We then calculate two $80 \times 80$ pairwise distance matrices before and after the temporal alignment. In Fig.~\ref{fig:gamblingdist}, we display these distance matrices before and after alignment,
and a histogram of relative changes in pairwise distances due to alignment (for all trajectories). 
From this figure, we see that the dynamic FCs generated by the Salience network are more homogenous 
compared with the ones generated by the CinguloParietal network.
This is reasonable because ROIs in the CinguloParietal network are actively 
involved in the gambling task \cite{10.3389/fpsyt.2016.00082}.  
The histogram in the last panel shows that there is a significant reduction ($\sim 10-20\%$) in distances
due to temporal alignment of trajectories. 

Next, these distance matrices are used as features for a binary classification task to study whether we can distinguish the dynamic FCs generated from different brain regions when people perform the same task.  The average classification rates before and after alignment using the DNN classifier are shown in Table.~\ref{tab:table0}. From this result, we can see that the alignment helps improve the classification rate for the class with higher variation.

\begin{table}[h!]
	\centering
	\caption{Average classification rates between FCs in the Salience and CinguloParietal networks before and after alignment.}
	\label{tab:table0}
	\begin{tabular}{lcc}
		\hline\hline
		&Salience & CinguloParietal \\
		
		\hline
		Before alignment  &80\% & 83\%  \\
		After alignment   &81\% & 91\%  \\
		\hline
	\end{tabular}
\end{table}

\subsubsection{Dynamic FCs under different tasks but for same ROIs}
Here we are interested in studying the classification of dynamic FCs under different tasks using the same set of ROIs. To save computational cost, in the following experiments we use $d_c$ for computing distances between trajectories. We present results from three scenarios: 
\begin{itemize}
	\item {\it Scenario 1: Classification and prediction based on ROIs in the VentralAttn and DorsalAttn networks}. \\
	We use 196 subjects from the HCP dataset. For four selected tasks - gambling, social, emotion and resting state - we construct dynamic FC trajectories for each subject using ROIs either in the Ventral or in the DorsalAttn networks. The VentralAttn network has 20 ROIs, resulting in 784, $20 \times 20 \times 20$ dynamic FC trajectories. The DorsalAttn network has 29 ROIs, resulting in 784, $29 \times 29 \times 20$ dynamic FC trajectories. We then calculate pairwise distances between 
	trajectories, with dimension reduction, and feed the distance features to the classifiers. 
	The average overall classification rates based on five-fold cross-valiation  are shown in Fig.~ \ref{fig:clf_vs_dim}.  There we can see that the classification rates are relatively stable, except when we reduce the dimension all the way to $d=5$. Notably, we can reduce the computational cost of comparing trajectories by a significant amount due to dimension-reduction and still retain good performance. In Fig.~\ref{fig:clf_vs_dim} we also present results obtained using trajectories directly as input for an SVM classifier, noted as SVM-t. Results show that using distances as features results in higher success rates in this low-dimensional classification task.
	
	A confusion matrix of classification results using all ROIs in DorsalAttn network and a DNN classifier is shown in Fig.~\ref{fig:clf_cm}. We see that for one case (resting state) the success rates reaches $100\%$, while it is lower for some cases.
	
	\begin{figure}
		\centering  
		\includegraphics[width=2.2in]{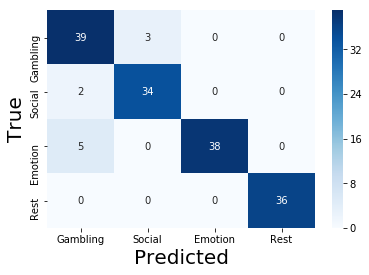}  
		\caption{Class-wised classification results in DosalAttn networks involving  four activities.} 
		\label{fig:clf_cm}
	\end{figure}
	
	\begin{figure}
		\centering  
		\includegraphics[width=2.5in]{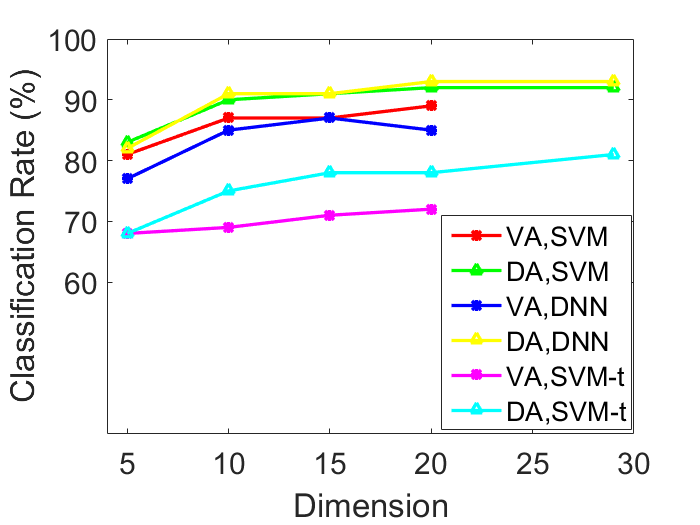}  
		\caption{Classification results in VentralAttn and DosalAttn networks involving 196 subjects and four activities with different dimensions. Top four lines were results using distance features as classifier input and bottom two lines were results using trajectories as input.}
		\label{fig:clf_vs_dim}
	\end{figure}

	\item Scenario 2: Classification of dynamic FC trajectories based on ROIs in other networks. \\ 
	Here we use 196 subjects and different sets of ROIs  -- ROIs in DorsalAttn (DA), RetrosplenialTemporal (RT), CinguloParietal (CP) and SMmouth (SM) networks) -- as the source for 
	data to be used in classification. Using the same settings as in Scenario 1, the average overall classification rates for the two classifiers are shown in Fig.~\ref{fig:hist_classification_networks}. 
	We study a four-class classification (gambling, social, emotion and resting state), versus a more difficult six-class classification (motor, gambling, social, language, emotion and resting state).
	We see that the success rates using the proposed metric reach $93\%$ in the four-class classification, but drop to $72\%$ in the six-class classification. In this setup, despite the simple architecture of given DNN classifier, its overall performance is as competitive as using SVM. One can seek performance improvement using more advanced neural network architectures. Here we also compare performance using the log-Euclidean metric between SPDMs for computing distances between covariance trajectories for comparison, represented as SVM-L in Fig.~\ref{fig:hist_classification_networks}. We see our proposed metric gives better overall performance, and log-Euclidean metric gives better performance in the six class classification problem using ROIs in DorsalAttn network. One can interpret that the two metrics are better at capturing different characteristcs of dynamic functional connectivity from different activities, so one can consider combining distance features calculated from different metrics for classifications in practice.

	\begin{figure}
		\centering  
		\includegraphics[width=3.8in]{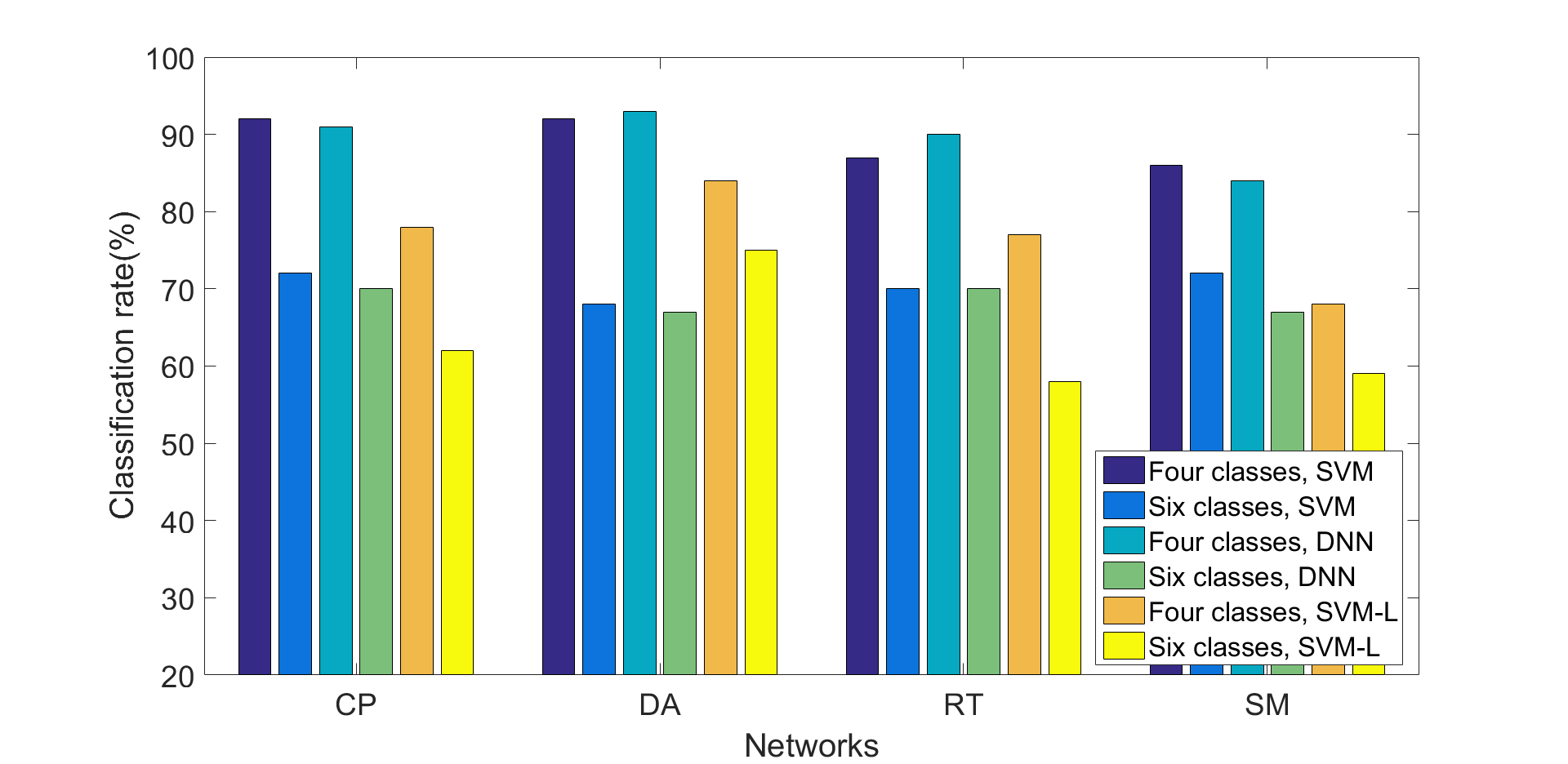}  
		\caption{Classification results from different networks using four activities and six activities.} 
		\label{fig:hist_classification_networks}
	\end{figure}

	\item Scenario 3: Classification of dynamic FCs with a large number of brain regions and dimension reduction with comparison. \\
	In this experiment, we involve all 333 ROIs  
	and four activities (gambling, social, emotion and resting-state) with 100 subjects for each activity. Here we also 
	compare performances of our proposed pipeline with the regular PCA method. To perform PCA, for each of the averaged ROI based multivariate time series in $\real^n$ ($n=333$ here), we first reduce its dimension to $\real^d$ using PCA, then transform it into covariance trajectory and calculate pairwise distances as earlier. We did not perform experiments in the case when $n=333$ because of the high computational and storage cost.
	
	The pairwise distance matrices between 400 trajectories with different degrees of dimension reduction are shown in Fig~\ref{fig:100subs_dr}. Despite different dimension reduction, the block patterns are still well preserved. Classification results using SVM classifier for these setups are presented in Table~\ref{tab:table10}.  We see that the proposed pipeline provides the highest classification rates across different dimensions, highlighting the power of the proposed method.

	\begin{figure}
		\centering  
		\begin{tabular}{ccc}
			\includegraphics[width=1.15in]{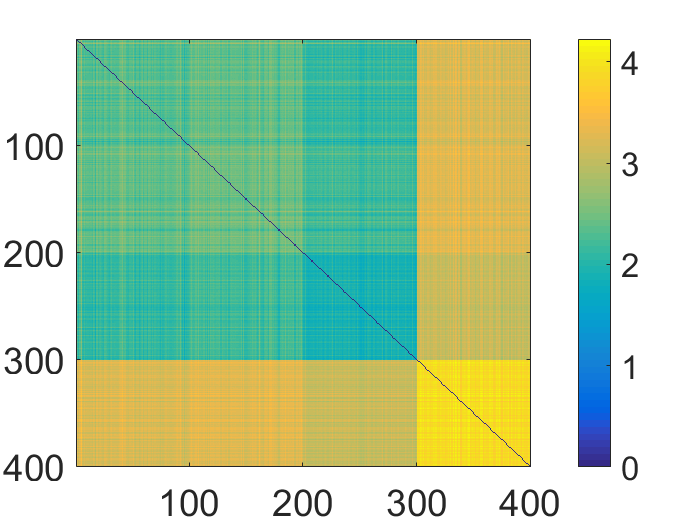}&
			\includegraphics[width=1.15in]{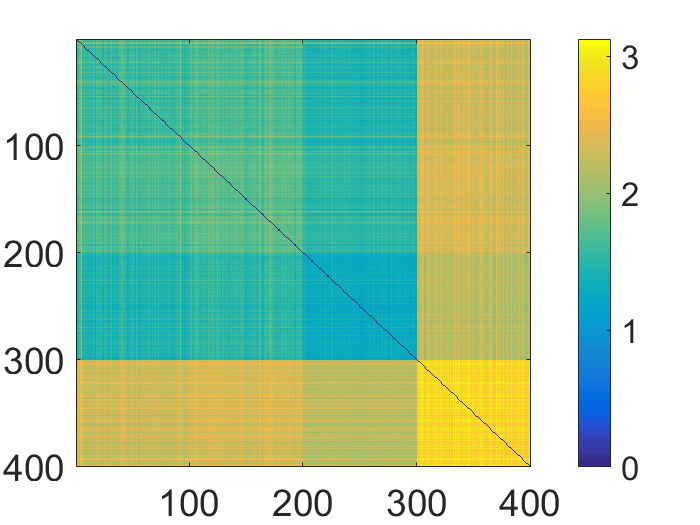}	&
			\includegraphics[width=1.15in]{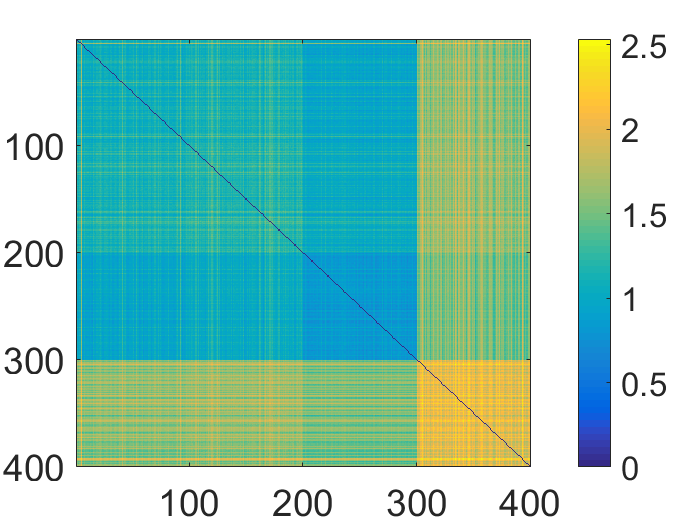}
			\\
			(a) & (b) & (c)\\
		\end{tabular}
		\caption{Pairwise distance matrix between 400 trajectories from four activities using 333 ROIs with dimension reduction. (a) (b) (c) show when $d = 100, 30, 10$ respectively.
		} \label{fig:100subs_dr}
	\end{figure}
	
	\begin{table}[h!]
		\tabcolsep 8pt
		\centering
		\caption{Classification rates of four activities with 333 ROIs with dimension reduction.}
		\label{tab:table10}
		\begin{tabular}{lcccc}
			\hline\hline
			DR method &d=10 & d=30 & d=100\\
			\hline
			Proposed & 80\% & 80\%& 82\% \\
			PCA      & 48\% & 59\%& 53\% \\
			\hline
		\end{tabular}
	\end{table}
	
	
\end{itemize}

Existing fMRI pattern classification literature utilizes data from different sources. Some works use particular datasets to classify mental disorders and some aim on classifying task/resting activities. To our knowledge, there are a few works using HCP dataset for classification of activities. One example is Zhang et al. \cite{ZHANGS2016}, where 
the classification accuracy is up to 100\% when classifying tfMRI/rsfMRI in a two-class classification problem. \cite{Chung2016ClassifyingHT} and \cite{7950674} also reach high classification rates 
in a two-task classification problem. In contrast, our results achieve high classification rates in a multi-task classification problem, and also reach 100\% classification rates for activities such as resting state. 

\begin{figure}[h!]
	\centering  
	\includegraphics[width=1.6in]{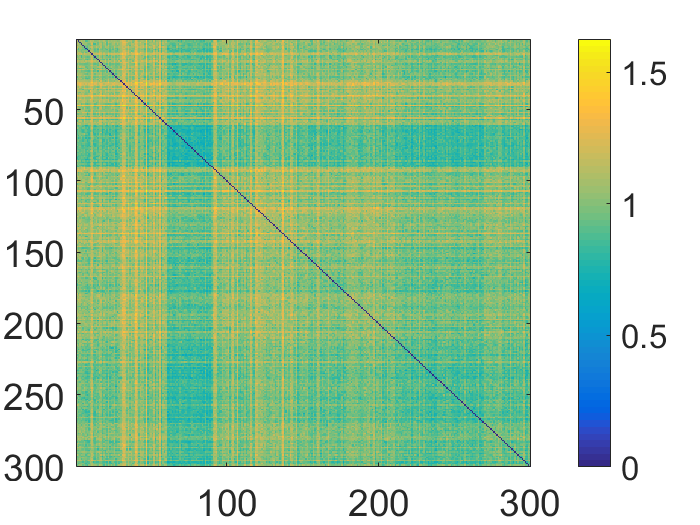}  
	\caption{Pairwise distances of 300 trajectories from 10 networks (FrontalParietal, SMhand, Salience, Visual, CinguloOperc, VentralAttn, DorsalAttn, RetrosplenialTemporal, CinguloParietal, SMmouth respectively) after 
		reducing dimensions to $d=4$.} 
	\label{fig:10networks}
\end{figure}

Another usage of the proposed framework is for 
comparing covariance trajectories with different dimensions, i.e. different number of 
ROIs.  In view of the dimension reduction used in our technique, one can bring 
different trajectories to the same dimension, thus making the comparison between theses trajectories possible. 
As an example, we use resting state data and different number of ROIs in 10 networks to form 
covariance trajectories with 30 subjects in each network, 
resulting on total of 300 trajectories. We project all covariance trajectories using $d = 4$ 
and compute  pairwise distances between all 300 trajectories. 
The results, shown  in Fig.~\ref{fig:10networks}, shows a strong consistent pattern in Salience network under resting state \cite{Salience2013}.

\section{Discussion}
In this paper we have presented a comprehensive  
framework for comparing multivariate fMRI time series, based on a distance on 
trajectories in the space SPDMs. 
We also derive a method for SPDM dimension reduction, 
which saves significant computational costs while preserving pairwise distances as much as possible. 
The method can be further used to compare covariance trajectories in different dimensions. 
Experimental results show high classification accuracies in multi-class classification problems, especially in emotion task and resting state data. 

We also notice that in a multi-class classification problem, distance features can be similar for certain tasks due to similar functional connectivity patterns of the selected ROIs, thus the resulting classification rates are relatively low in these tasks. 
In terms of classification, we expect that given the distance features, using advanced deep learning frameworks and ensemble models may improve performance.

\section*{Acknowledgements}

\begin{appendices}
	\section*{Appendix}
	\subsection {Discussion on pseudoinverse of $\hat{P}$} 
	For an $n \times d$ orthogonal matrix $B$, an $n \times n$ SPDM $P$ and 
	a projection $Q = B^T P B \in \real ^{d \times d}$, the reconstruction of $P$ is given by: 
	$\hat{P}=BQB^T$, such that $B^TB =I_d$.
	Since $\hat{P}$ is singular, we cannot invert it, but can use the notion of 
	a pseudoinverse instead. 
	Let the SVD of $Q$ be $Q=U Q_0 U^T$, and an SVD of $\hat{P}$ be 
	$\hat{P} =
	\begin{bmatrix} 
	U_1 & U_2
	\end{bmatrix}
	\quad
	\begin{bmatrix} 
	S_1 & 0 \\
	0 & S_2 
	\end{bmatrix}
	\quad
	\begin{bmatrix} 
	U_1 & U_2 
	\end{bmatrix}^T
	$. Here $S_1 \in \real^{d \times d}$ and and $S_2 \in \real^{(n-d) \times (n-d)}$. 
	Due to $\hat{P}$ not being full ranked,  $S_2$ is a matrix of zeros, and 
	$\hat{P} = U_1 S_1 {U_1}^T$. 
	Its pseudo inverse is thus defined to be
	$\hat{P}^{-} = U_1 {S_1}^{-1} {U_1}^T$.
	
	On the other hand,
	$\hat{P}=BQB^{T}=BUQ_0U^TB^T$. 
	Since $U$ is a rotation matrix, $BU$ is just a rotation of $B$, and $BU$ is still orthonormal. Setting 
	$BU=U_1$ and $Q_0=S_1$, we get SVD of $\hat{P}$, and its pseudo inverse can be written as 
	$\hat{P}^{-} = BU{Q_0}^{-1}(BU)^T = B Q^{-1} B^T$.
	
	\subsection {Proof of Lemma~\ref{lemma:lm1}}
	Given $\hat{P} = BQB^T$ and $\hat{P}^{-} = B Q^{-1} B^T$,  we have 
	$$\hat{P_i}^{-} \hat{P_j}^{2} \hat{P_i}^{-}
	=(B{Q_i}^{-1} B^{T}) (BQ_j B^{T}) (BQ_j B^{T}) (B{Q_i}^{-1} B^{T})\ ,
	$$
	for all $i$ and $j$.
	Since $B^{T}B=I_d$, this equals 
	$$B{Q_i}^{-1} B^{T} BQ_j B^{T} BQ_j B^{T} B{Q_i}^{-1} B^{T} \ ,
	$$ 
	which equals
	$B{Q_i}^{-1}Q_j^{2}{Q_i}^{-1}B^{T}
	= B{Q_{ij}}B^{T}$. 
	Therefore, we can write
	$$
	\|{P_i}^{-}{P_j}^{2}{P_i}^{-} - \hat{P_i}^{-}\hat{P_j}^{2}\hat{P_i}^{-} \| =
	\| P_{ij} - B{Q_{ij}}B^T \|\ .
	$$
	
	\subsection{Proof of Lemma~\ref{lemma:lm2}}
	Starting with the left side and simplifying, we get: 
	\[ \begin{array}{lcl}
	\argmin_{B,Q_{ij}} \sum_{i,j} \| P_{ij} - B Q_{ij} B^T \|^2\ \\
	= \argmin_{B,Q_{ij}}  \sum_{i,j}  tr( P_{ij} - B Q_{ij} { B}^T )(P_{ij} -  { B} Q_{ij} {B}^T )\\
	=\argmin_{{ B},Q_{ij}}  \sum_{i,j}  tr(P_{ij} P_{ij}- P_{ij} { B} Q_{ij} { B}^T \\
	- { B} Q_{ij} { B}^T P_{ij}+{ B} Q_{ij} { B}^T { B} Q_{ij} { B}^T)\\
	=\argmin_{{ B},Q_{ij}}  \sum_{i,j}  tr(P_{ij} P_{ij}-2 { B}^T P_{ij} { B} Q_{ij} + Q_{ij} Q_{ij})\ .
	\end{array} \]
	To simplify notation,  
	we replace the trace operator with the matrix inner product, i.e 
	$tr({ A} { B}) = \left<{  A}, { B} \right> = \sum_{i,j} a_{ij}b_{ij}$. Now, for each $i$, we have
	\[ \begin{array} {lcl}
	tr( P_{ij} P_{ij} - 2 { B}^T P_{ij} { B} Q_{ij} + Q_{ij} Q_{ij}) \\
	=  \argmin_{{B},Q_{ij}} (\left< P_{ij}, P_{ij} \right> - 2\left< { B}^T P_{ij} { B}, Q_{ij} \right> + \left< Q_{ij}, Q_{ij} \right>) \\
	=  \argmin_{{ B},Q_{ij}} ( - \left< { B}^T P_{ij} { B},{ B}^T P_{ij} { B} \right>  \\
	+ \left<  { B}^T P_{ij} { B} - Q_{ij},  { B}^T P_{ij} { B} - Q_{ij} \right>)\ .
	\end{array} \]
	
	It is easy to see that, to minimize the objective function, we need to set ${ B}^T P_{ij} { B} = Q_{ij}$ for each $i,j$.  By substituting  $ Q_{ij} = { B}^T P_{ij} { B} $, we have
	\[ \begin{array} {lcl}
	\argmin_{{ B},Q_{ij}} tr( P_{ij} P_{ij} - 2 { B}^T P_{ij} { B} Q_{ij} + Q_{ij} Q_{ij}) \\
	= \argmin_{{ B}} - \left< { B}^T P_{ij} { B},
	{ B}^T P_{ij} { B} \right> \\
	= \argmax_{{ B}}  tr( { B}^T P_{ij} { B}{ B}^T P_{ij} { B}) \ .
	\end{array} \]
	Therefore, for all $i,j$s, 
	\[ \begin{array}{lcl}
	\argmin_{{ B},Q_{ij}} \sum_{i,j} \| P_{ij} -  { B} Q_{ij} { B}^T \|_F^2\ \\
	= \sum_{i,j} \argmax_{{ B}}  tr( { B}^T P_{ij} { B}{ B}^T P_{ij} { B}) \ . \Box
	\end{array} \]
\end{appendices}

\bibliographystyle{IEEEtran}
\bibliography{FCT,bibfile}

\end{document}